\documentclass[10pt,twocolumn,letterpaper]{article}

\usepackage{bibspacing}
\usepackage{cvpr}
\usepackage{times}
\usepackage{epsfig}
\usepackage{graphicx}
\usepackage{amsmath}
\usepackage{amsmath,bm}
\usepackage{amssymb}
\usepackage{indentfirst}
\usepackage{times}
\usepackage{cite}
\usepackage{multirow}
\usepackage{diagbox}
\usepackage{array}
\usepackage{color}
\usepackage{float}
\usepackage{caption}
\usepackage{algorithm}
\usepackage{algpseudocode}
\usepackage{amsmath}
\usepackage{graphics}
\usepackage{epsfig}
\usepackage{mathrsfs}
\usepackage{subfigure}
\usepackage{verbatim}
\usepackage{amsfonts,amssymb}
\usepackage{makecell}
\usepackage{booktabs}
\usepackage[pagebackref=true,breaklinks=true,letterpaper=true,colorlinks,bookmarks=false]{hyperref}

\newcolumntype{I}{!{\vrule width 1pt}}
\newlength\savewidth
\newcommand\shline{\noalign{\global\savewidth\arrayrulewidth
                            \global\arrayrulewidth 1pt}%
                   \hline
                   \noalign{\global\arrayrulewidth\savewidth}}

 \cvprfinalcopy 

\def\cvprPaperID{2574} 

\ifcvprfinal\fi
\begin{document}


\title{\textbf{HashGAN:Attention-aware Deep Adversarial Hashing \\for Cross Modal Retrieval}}

\author{Xi Zhang$^{~1}$~~ Siyu Zhou$^{~1}$~~ Jiashi Feng$^{~2}$~~ Hanjiang Lai$^{~1}$\\
        Bo Li$^{~1}$~~ Yan Pan$^{~1}$~~ Jian Yin$^{~1}$~~ Shuicheng Yan$^{~2,3}$\\
$^{1~}$Sun Yat-Sen University, China\\
$^{2~}$National University of Singapore, Singapore\\
$^{3~}$QiHoo 360 Artificial intelligence institute, China\\
}
\date{}
\maketitle

\begin{abstract}
As the rapid growth of multi-modal data, hashing methods for cross-modal retrieval have received considerable attention. Deep-networks-based cross-modal hashing methods are appealing as they can integrate feature learning and hash coding into end-to-end trainable frameworks. However, it is still challenging to find content similarities between different modalities of data due to the heterogeneity gap. To further address this problem, we propose an adversarial hashing network with attention mechanism to enhance the measurement of content similarities by selectively focusing on informative parts of multi-modal data. The proposed new adversarial network, HashGAN, consists of three building blocks: 1) the feature learning module to obtain feature representations, 2) the generative attention module to generate an attention mask, which is used to obtain the attended (foreground) and the unattended (background) feature representations, 3) the discriminative hash coding module to learn hash functions that preserve the similarities between different modalities. In our framework, the generative module and the discriminative module are trained in an adversarial way: the generator is learned to make the discriminator cannot preserve the similarities of multi-modal data w.r.t. the background feature representations, while the discriminator aims to preserve the similarities of multi-modal data w.r.t. both the foreground and the background feature representations.
Extensive evaluations on several benchmark datasets demonstrate that the proposed HashGAN brings substantial improvements over other state-of-the-art cross-modal hashing methods.
\end{abstract}

\section{Introduction}
\label{introduction}
\begin{figure}[htb]
\centering
\includegraphics[width=0.48\textwidth]{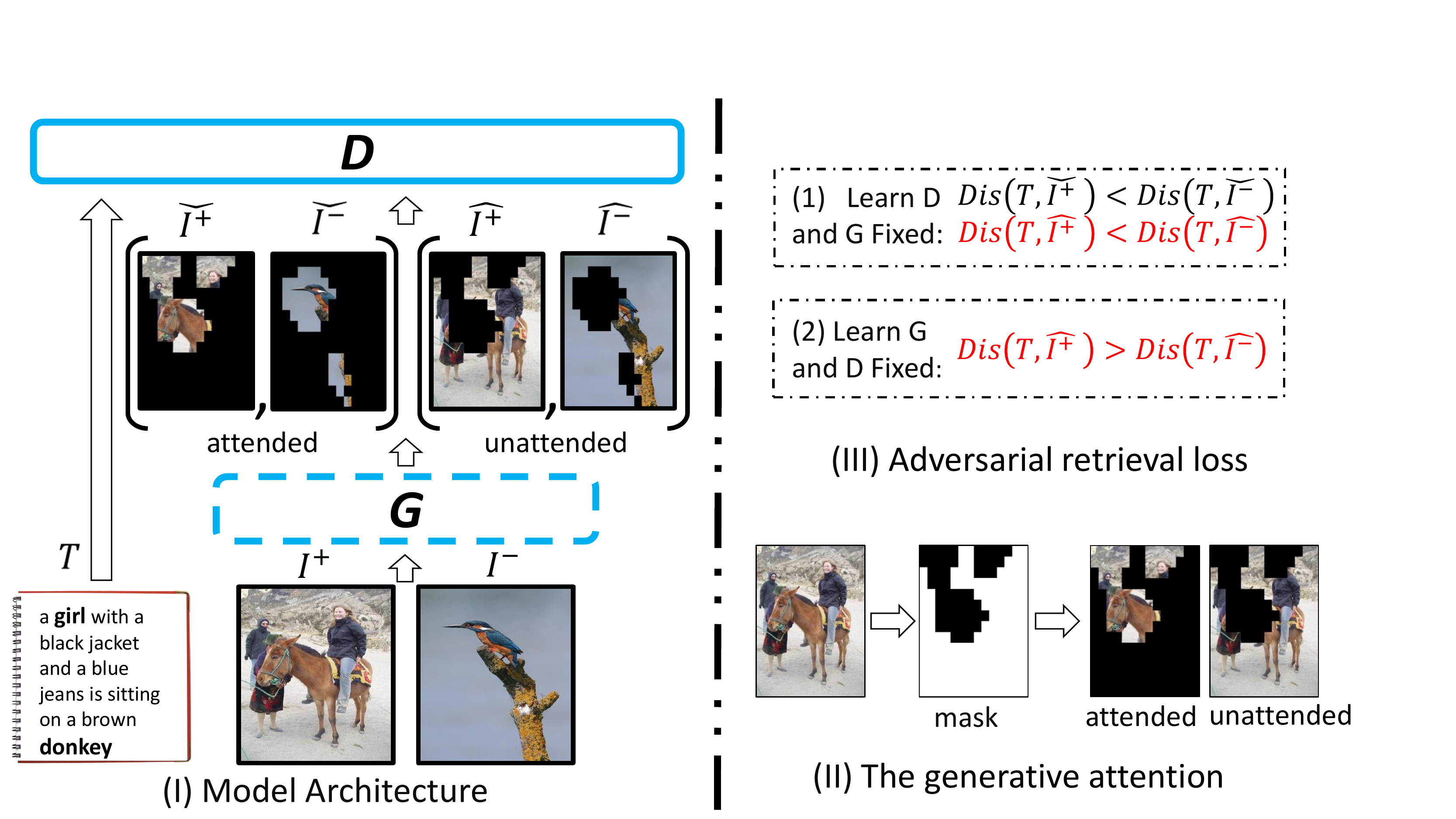}
\setlength{\abovecaptionskip}{5pt}\caption{Main idea of HashGAN. Take $Text{\rightarrow}Image$ task as an example, the $T$ and $I^+$ are similar while $T$ and $I^-$ are irrelevant, and $G$ denotes as the generator for generating attention masks while $D$ is our desired similarity-preserving hash functions. The two images $I^+$ and $I^-$ go through the generator $G$, which divides the data into attended/foreground samples \{$\breve{I}^+$, $\breve{I}^-$\} and unattended/background samples \{$\hat{I}^+$, $\hat{I}^-$\}. The process of generative attention module is shown in (II). Finally, these four images and the query are fed into the discriminator $D$.
We train on discriminator and generator in an adversarial way (III): (1) the discriminator aims to learn the hash functions that preserve the similarities for both the foreground samples and the background samples, (2) the generator aims to generate attention masks that make discriminator cannot preserve the similarities of the background samples.
}
\label{figure:example}
\end{figure}

\begin{figure*}[t]
\centering
\includegraphics[width=0.9\textwidth]{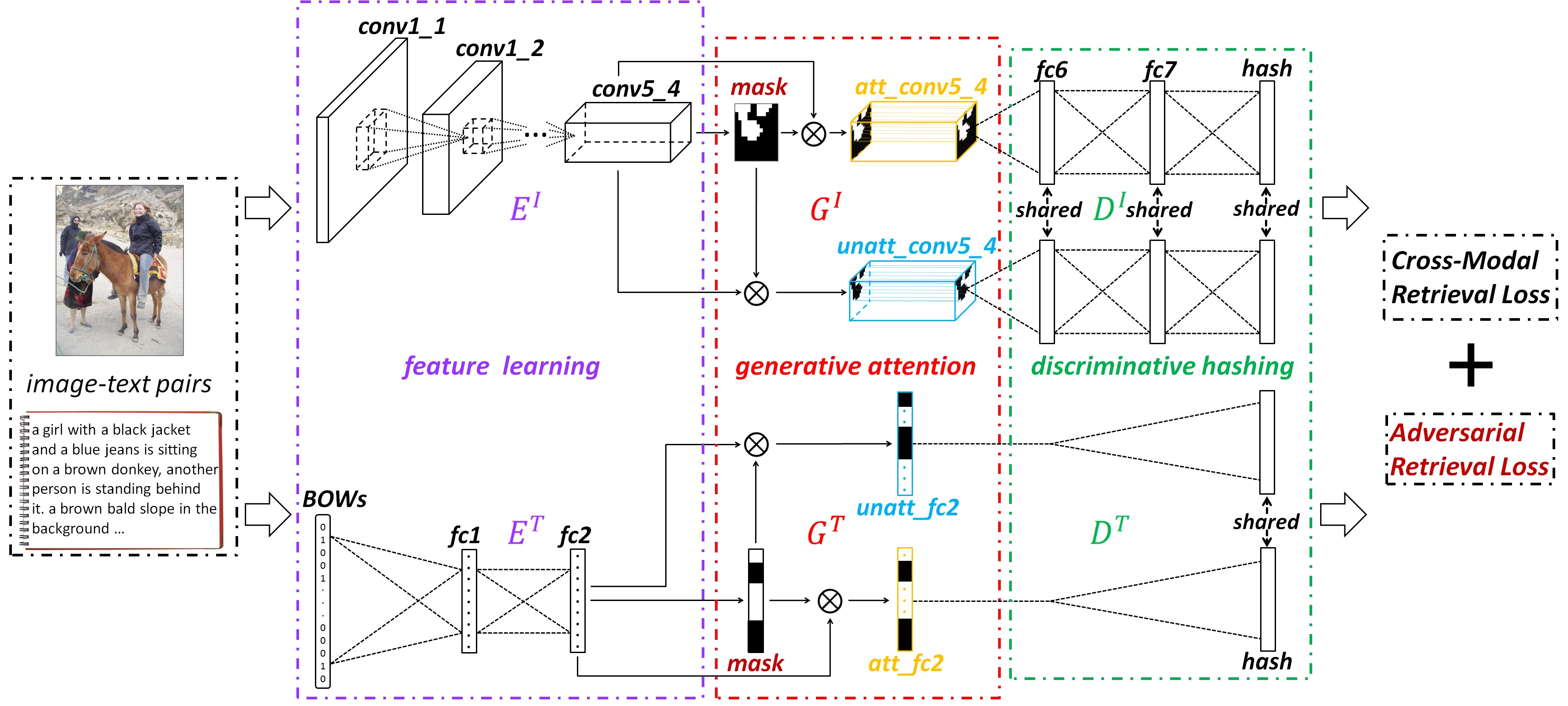}
\setlength{\abovecaptionskip}{5pt}\caption{Overview of HashGAN. Above is image modality branch, and below is text modality branch. Each branch is divided into three parts: feature learning ($E^I$ and $E^T$), generative attention ($G^I$ and $G^T$) and discriminative hashing ($D^I$ and $D^T$). The feature learnings map the input multi-modal data into high-level features representations. Then the generators learn the attention masks for these features representations. The attended (foreground) features and the unattended (background) features are generated via the attention masks. Finally, discriminators encode all features into binary codes and learn similarity-preserved hash functions. We train the discriminator and generator alternately, which generators maximize the retrieval loss of background features for generating good masks while discriminators minimize the error for both foreground features and background features for obtaining efficient binary codes. }
\label{figure:overview}
\end{figure*}

Due to the fast development of Internet, different types of media data grow rapidly, e.g., texts, images and videos. These different types of data may describe the same events or topics. For example, the photos in Flickr are allowed users to give interactive comments. Hence, developing a retrieval model for multi-modal data is a desired requirement. Cross-modal retrieval, which takes one type of data as the query and return the relevant data of another type, is receiving increasing attention since it is a natural searching way for multi-modal data. The solution methods can be roughly divided into two categories~\cite{wang2016comprehensive}: real-valued representation learning and  binary representation learning. Since the low storage cost and fast retrieval speed of the binary representation, we only focus on cross-modal binary representation learning (i.e., Hashing) in this paper.

To date, various cross-modal hashing algorithms~\cite{zhen2012co,zhang2014large,ding2014collective,SePH,DCMH,cao2016deep,yang2017pairwise} have been proposed for embedding correlations among different modalities of data. In the cross-modal hashing procedure, feature extraction is considered as the first step for representing all modalities of data, and then one project these multi-modal features into a common Hamming space for future search. Many methods~\cite{ding2014collective,zhang2014large} use shallow architecture for feature extraction. For example, collective matrix factorization hashing (CMFH)~\cite{ding2014collective} and semantic correlation maximization (SCM)~\cite{zhang2014large} use the hand-crafted features. Recently, deep learning has also been adopted for cross-modal hashing due to its powerful ability of learning good representations of data. The representative works of deep-network-based cross-modal hashing includes deep cross-modal hashing (DCMH)~\cite{DCMH}, deep visual-semantic hashing (DVSH)~\cite{cao2016deep}, pairwise relationship guided deep hashing (PRDH)~\cite{yang2017pairwise} and so on.

In parallel, the computational model of ``attention" has drawn much interest due to its impressive result in various applications, e.g., image caption~\cite{xu2015show}. It is also desired for cross-modal retrieval problem. For example, as shown in Figure~\ref{figure:example}, given a query \textit{girl sits on donkey}, if we can locate the more informative objects in image (e.g., the black regions), the more accuracy can be obtained. To the best of our knowledge, the attention mechanism has not been well explored for cross-modal hashing.

In this paper, we propose an adversarial hashing network with attention mechanism for cross-modal hashing. Ideally, good attention masks should locate discriminative regions, which also mean the unattended regions of data are uninformative and hard to preserve similarities. Hence, in our proposed network, adaptive attention masks are generated for the multi-modal data, then the learned masks divide the data into attended samples(only keep foregrounds of the data) and unattended samples(only keep backgrounds of the data).  Hinging on such attention masks, a good discriminative hashing should preserve the similarities for both the foreground samples (which can be viewed as easy examples) and background samples (hard examples) for enhancing the robustness and performance of the learned hash functions. And the good generator should generate attention masks that make discriminator cannot preserve the similarities of the background samples, for unattended regions of data should not be discriminative.

Based on this, we present a new adversarial model called HashGAN, which is illustrated in Figure~\ref{figure:overview} and consists of three major components: (1) feature learning module which uses CNN or MLP to extract high level semantic representations for the multi-modal data, (2) generative attention module which generates the adaptive attention masks and divides the feature representations into the attended and the unattended feature representations, and (3) discriminative hashing module which focus on learning the binary codes for the multi-modal data. HashGAN trains two adversarial networks alternatively: the discriminator is learned to preserve the similarities for both the easy foreground feature representations and the hard background feature representations, while the generator learns to produce masks that make the discriminator fails to keep similarities of the background feature representation. The adversarial retrieval loss and cross-modal retrieval loss are proposed to obtain good attention masks and powerful hash functions.

The main contributions of our work are three-fold. First, we propose an attention-aware method for cross-modal hashing problem, which is able to detect the informative regions of multi-modal data. Second, we propose an HashGAN for learning effective attention masks and compact binary codes simultaneously. Third, we quantitatively evaluate the usefulness of attention in cross-modal hashing and our method yields better performances by comparing with several state-of-the-art methods.

\section{Related Work}
\label{related_work}
\subsection{Cross-modal Hashing}
According to the utilized information for learning the common representations, cross-modal hashing can be categorized into three groups~\cite{wang2016comprehensive}: 1) unsupervised methods~\cite{wang2015learning}, 2) pairwise based methods~\cite{masci2014multimodal,zhen2012co} and 3) supervised methods~\cite{yu2014discriminative,cao2016correlation}. The unsupervised methods only use co-occurrence information to learn hash functions for multi-modal data. For instance, cross-view hashing (CVH)~\cite{sun2008least} extends spectral hashing from uni-modal to multi-modal scenario. The pairwise based methods use both the co-occurrence information and similar/dissimilar pairs to learn the hash functions. Bronstein et al.~\cite{he2011maximum} proposed cross-modal similarity sensitive hashing (CMSSH), which learn the hash functions to ensure that if two samples (with different modalities) are relevant/irrelevant, their corresponding binary codes are similar/dissimilar. The supervised methods exploit label information to learn more discriminative common representation. Semantic correlation maximization (SCM)~\cite{zhang2014large} uses label vector to obtain the similarity matrix and reconstruct it through the binary codes.

However, most of these works are based on hand-crafted features. Recently, deep learning methods show that they can effectively discover the correlations across different modalities. The most representative work is deep cross-modal hashing (DCMH)~\cite{DCMH}. DCMH integrates feature learning and hash-code learning into the same framework. Cao et.al.~\cite{cao2016deep} proposed deep visual-semantic hashing (DVSH), which utilizes both the convolutional neural network (CNN) and long short term memory (LSTM) to separately learn the common representations for each modality. Pairwise relationship guided deep hashing (PRDH)~\cite{yang2017pairwise} also adopts deep CNN models to learn feature representations and hash codes simultaneously.

However, all these methods encode an entire data point into a binary representation. Few works attend to introduce attention mechanism into cross-modal hashing.
\subsection{Attention Models}
Attention-aware methods capture where the model should focus on when performing a particular task. The attention mechanism has been proved to be very powerful in many applications, such as image classification~\cite{ba2015}, image caption~\cite{xu2015show}, image question answering~\cite{yang2016stacked}, video action recognition~\cite{sharma2015action} and etc. For example, Xu et al.~\cite{xu2015show} proposed two forms of attention for image caption: a ``hard" attention mechanism trained by REINFORCE and a ``soft" attention mechanism trained by standard back-propagation methods. Stacked attention networks (SANs)~\cite{yang2016stacked} take multiple steps to progressively focus the attention on the relevant regions and lead to a better answer for image QA. Sharma et al.~\cite{sharma2015action} proposed a soft attention based model for action recognition which uses recurrent neural networks (RNNs) with long short-term memory (LSTM) unit to obtain both the spatial and temporal information.
\subsection{Generative Adversarial Network}
Generative adversarial networks (GANs) have been received a lot of interest in generative modelling problems. The original GAN~\cite{goodfellow2014generative} train two models: the discriminative model $D$ and the generative model $G$. The discriminative model learns to determine whether a sample is from the model distribution or data distribution. The generative model attempts to produce a sample that can fake the discriminative model.

Recently, several approaches have been proposed to improve the original GAN. For example, DCGAN~\cite{radford2015unsupervised}, CGAN~\cite{mirza2014conditional} and Wasserstein GAN~\cite{arjovsky2017wasserstein}. IRGAN~\cite{wang2017irgan} is a recently proposed method for information retrieval, in which the generative retrieval focusing on predicting relevant documents and the discriminative retrieval focusing on predicting relevancy given a query document pair. Different from our method, IRGAN is designed for uni-modal retrieval and it is not an attention-aware method, yet.

In this paper, we extend GAN to cross-modal hashing. We carefully design a new GAN, called HashGAN, to generate attention-aware common representations and to learn similarity-preserve hash functions.

\section{HashGAN}
\label{our_method}

\subsection{Problem Definition}
Suppose there are \(n\) training samples, each of which is represented in several modalities, e.g., audio, video, image, text, etc. In this paper, we only focus on two modalities: text and image. Note that our method can be easily extended to other modalities.

We denote the training data as $\{I_i,T_i \}_{i=1}^n$, where $I_i$ is the $i$-th image and $T_i$ is the corresponding text description of image $I_i$. We also have a cross-modal similarity matrix $S$, where $S(i,j) = 1$ means the $i$-th image and the $j$-th text are similar and $S(i,j) = 0$ means they are dissimilar.

The goal of cross-modal hashing is to learn two mapping functions to transform image and text into a common binary codes space respectively, in which the similarities between paired images and texts are preserved. Formally, Let $H^I \in \{0,1\}^q$ and $H^T \in \{0,1\}^q$ be denoted as the generated $q$-bit binary codes for image and text, respectively. If $S(i,j) = 1$, the $i$-th image and the $j$-th text are similar. Hence, the Hamming distance between $H^T_j$ and $H^I_i$ should be small. When $S(i,j) = 0$, the Hamming distance between them should be large.

\subsection{Network Architecture}
We propose HashGAN for cross-modal problem, which contains three type of networks: 1) feature learning networks for obtaining the high-level representations of the multi-modal data, 2) generative attention network for learning the attention distributions, and 3) discriminative hashing networks for learning the binary codes for cross-modal image retrieval.

\subsubsection{Feature Learning Components: $E^I$ and $E^T$}
\begin{figure}[h]
\centering
\includegraphics[width=0.45\textwidth]{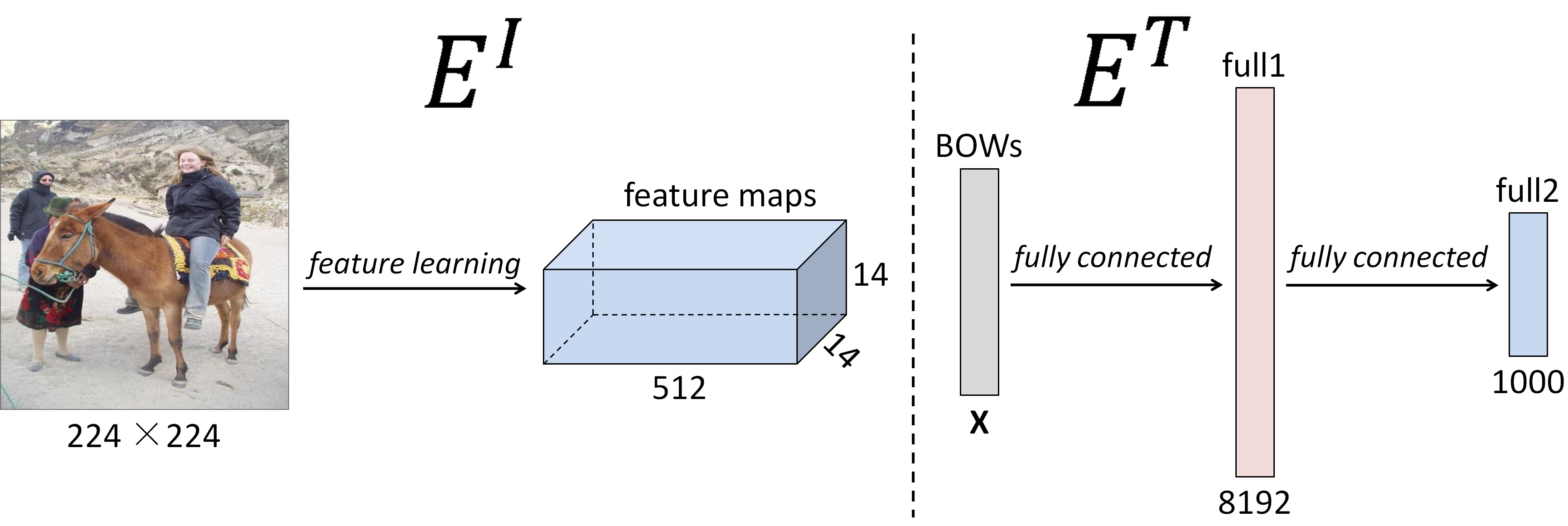}
\setlength{\abovecaptionskip}{5pt}\caption{Feature learning module for image modality $E^I$ and text modality $E^T$.}
\label{figure:feature-learning}
\end{figure}

For image modality, the convolutional neural network is used to obtain the high-level representation of images. In this paper, we use VGGNet~\cite{simonyan2014very} as the basic network to generate the feature maps as shown in Figure~\ref{figure:feature-learning}. Let $f^{I}_i = E^I(I_i)$ be denoted as the image feature maps from the $i$-th raw image.

For text modality, we use multi-layer perceptron (MLP) to obtain the powerful semantic representation of texts. Following DCMH~\cite{DCMH}, we also use bag-of-words (BOW) as the feature representation for text modality. There are two fully-connected layers as shown in Figure~\ref{figure:feature-learning}.  We denote $f^{T}_i = E^T(T_i)$ as the feature vector for the $i$-th text.

\subsubsection{Generative Attention Components: $G^I$ and $G^T$}
\label{attention_networks}
With the powerful image feature maps $f^{I}$ and the text feature vector $f^{T}$, we first feed them into one layer neural network, i.e., a convolutional layer with $1 \times 1$ kernel size for image feature maps and a fully-connected layer for text feature vector, and then followed by a softmax function and a threshold function to generate the attention distribution over the regions of multi-modal data.

\begin{figure}[t]
\centering
\includegraphics[width=0.35\textwidth]{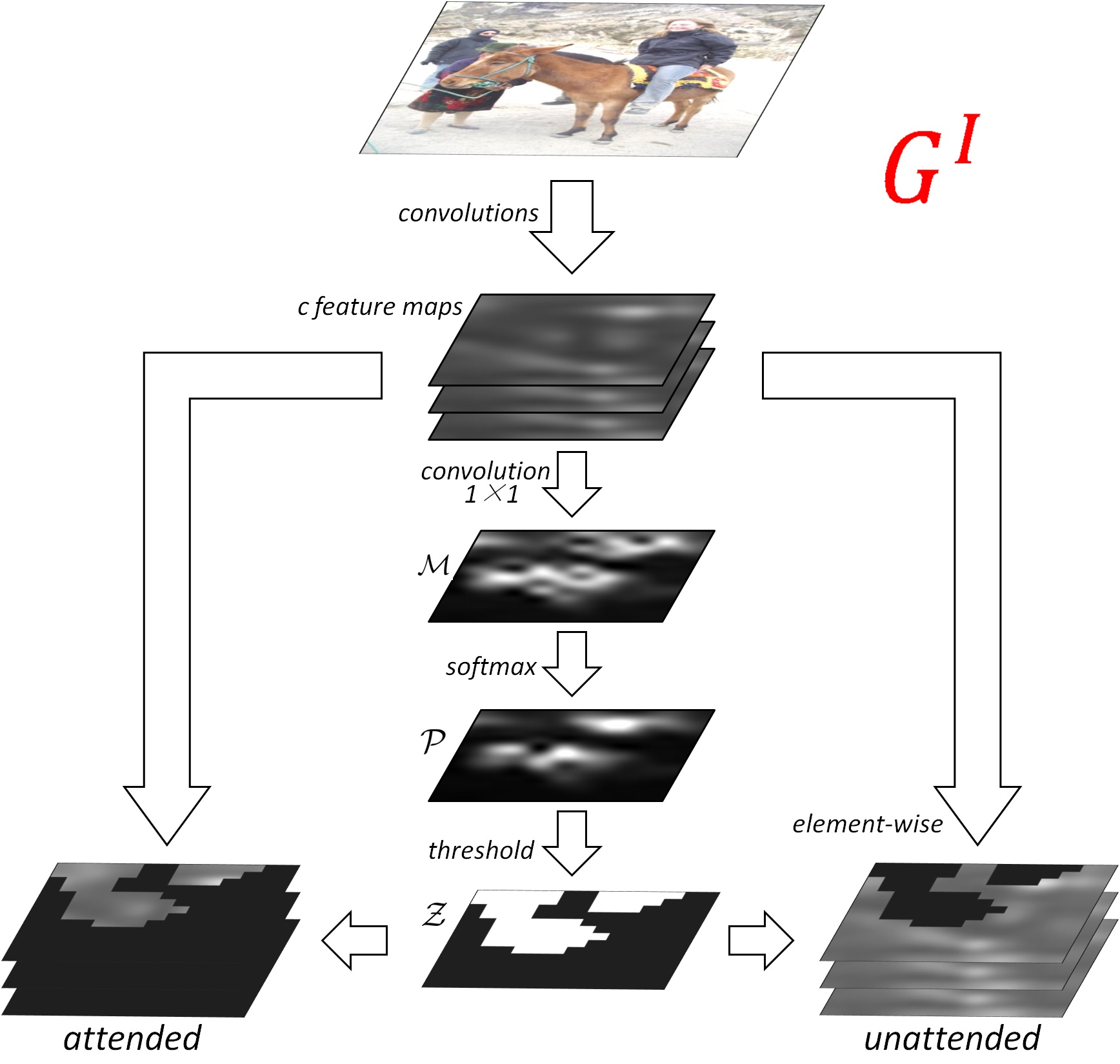}
\setlength{\abovecaptionskip}{5pt}\caption{Binary mask generated by $G^I$ in the image branch.}
\label{figure:imagemask}
\end{figure}

More specially, Figure~\ref{figure:imagemask} shows the pipeline in details for processing image modality. Suppose $f^I_i = E^I(I_i) \in \mathbb{R}^{H \times W \times C}$ is the feature maps for the $i$-th image, where $H$, $W$ and $C$ are the height, weight and channels of the feature maps, respectively. In the first step, the feature maps are mapped into the mask $m_i^I \in \mathbb{R}^{H \times W \times 1}$ by a convolutional layer with $1\times1$ kernel size. Next, the mask $m_i^I$ goes through a \emph{softmax} layer and the output is denoted as $p_i^I$, which is defined as
\begin{equation}
\centering
p_i^I(h,w)= \frac{e^{m_i^I(h,w)}}{\sum_{h=1}^{H}\sum_{w=1}^{W}{e^{m_i^I(h,w)}}},
\label{equ:softmax}
\end{equation}
where $m_i^I(h,w)$ and $p_i^I(h,w)$ denote the value in the $h$-th row and $w$-th column of the matrix $m_i^I$ and the matrix $p_i^I$, respectively. The elements in $p_i^I$ form a probability distribution, where $p_i^I(h,w)>0$ and $\sum{p_i^I(h,w)}=1$.

A larger value in $p^I$ correspond to the foregrounds and the backgrounds may have a smaller response. Thus, in the third step, we add a \emph{threshold} layer to divide the data into the attended regions and the unattended regions, which is defined as
\begin{equation}
\centering
z_i^I(h,w)=
\begin{cases}
1~~~~~~~~p_i^I(h,w)~\ge~\alpha\\
0~~~~~~~~p_i^I(h,w)~<~\alpha
\end{cases}
\label{equ:threshold}
\end{equation}
where $\alpha$ is a predefined threshold. We set $\alpha=\frac{1}{H{\times}W}$ in our experiment. The output of the threshold layer is a binary mask, with the elements inside be either 0 or 1. The regions with the value 1 are regarded as the foregrounds or the regions that are attend to, while other regions are regarded as background regions.

Based on the attention distribution, we can calculate the attention-aware feature maps and inattention-aware feature maps for the $i$-th image by multiplying the binary mask in element-wise, which is formulated as
\begin{equation}
\begin{aligned}
&\breve{f}^I_i (h,w,c) = z^I_i(h,w) f^I_i (h,w,c),& \\
&\hat{f}^I_i (h,w,c) = (1 - z^I_i(h,w)) f^I_i (h,w,c),&\\
\end{aligned}
\end{equation}
for all $h, w$ and $c$. The foreground is $\breve{f}^I_i$ and the background is $\hat{f}^I_i$. For ease of representation, we denote the whole procedures as $[\breve{f}^I_i, \hat{f}^I_i ]= G^I (f^I_i)$.

\begin{figure}[t]
\centering
\includegraphics[width=0.4\textwidth]{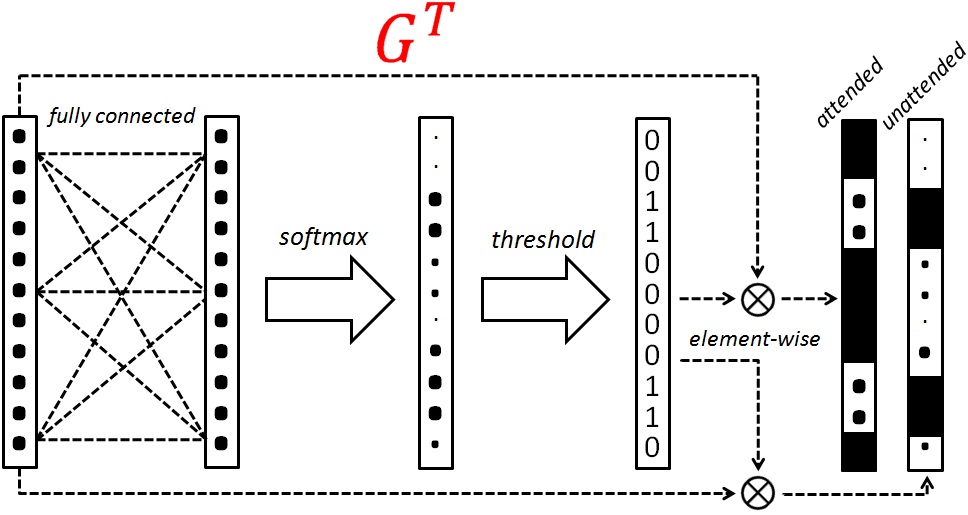}
\setlength{\abovecaptionskip}{5pt}\caption{Binary mask generated by $G^T$ in the textual branch.}
\label{figure:textmask}
\end{figure}

For text modality, we imitate the pipeline similar to image modality which is shown in Figure~\ref{figure:textmask}. Since there are feature vectors rather than feature maps, we use fully-connected layer instead of the convolutional layer, then it is fed to \emph{softmax} and \emph{threshold} respectively. Formally, we compute
\begin{equation}
\begin{aligned}
&m^T_i = \text{relu}(W^T f^{T}_i + b^T),& \\
&p^T_i = \text{softmax}(m^T_i),&\\
&z^T_i = \text{threshold}(p^T_i),& \\
&\breve{f}^T_i = z^T_i \otimes f^T_i,& \\
&\hat{f}^T_i = (1 - z^T_i) \otimes f^T_i,&
\end{aligned}
\end{equation}
where $W$ and $b$ are two parameters in the fully-connected layer, and $\otimes$ is Kronecker product. We denote $[\breve{f}^T_i,\hat{f}^T_i] = G^T(f^T_i)$ as the attention-aware and inattention-aware features for $i$-th text.

\par
While taking the derivative of the \emph{threshold} function directly is incompatible with the back-propagation in training. Specifically, suppose that $\mathcal{F}$ is the loss function, we need to use $\frac{{\partial} \mathcal{F} }{{\partial}p}$ in updating the network parameters by stochastic gradient descent (SGD) during back-propagation. However, the derivative $\frac{{\partial}z}{{\partial}p}$ in the \emph{threshold} layer is almost zero everywhere according to the definition of $z$. Besides, by the chain rule: $\frac{{\partial} \mathcal{F} }{{\partial}p}=\frac{{\partial} \mathcal{F} }{{\partial}z}\cdot\frac{{\partial}z}{{\partial}p}$, we can see that $\frac{{\partial} \mathcal{F} }{{\partial}p}$ is also nearly zero immediately. Eventually, such an almost zero-valued node may block the back-propagation process.
\par
To address this issue, we follow the idea proposed in \cite{derivative}, which uses the straight-through estimator to estimate or propagate the gradients of the \emph{threshold} function. That is to say, we ignore the derivative $\frac{{\partial}z}{{\partial}p}$, and set $\frac{{\partial} \mathcal{F} }{{\partial}p}$ by $\frac{{\partial} \mathcal{F} }{{\partial}z}$ as an estimator.

\subsubsection{Discriminative Hashing Components: $D^I$ and $D^T$}
The discriminator networks encode the high-level features for two modalities into binary codes.

\begin{figure}[h]
\centering
\includegraphics[width=0.45\textwidth]{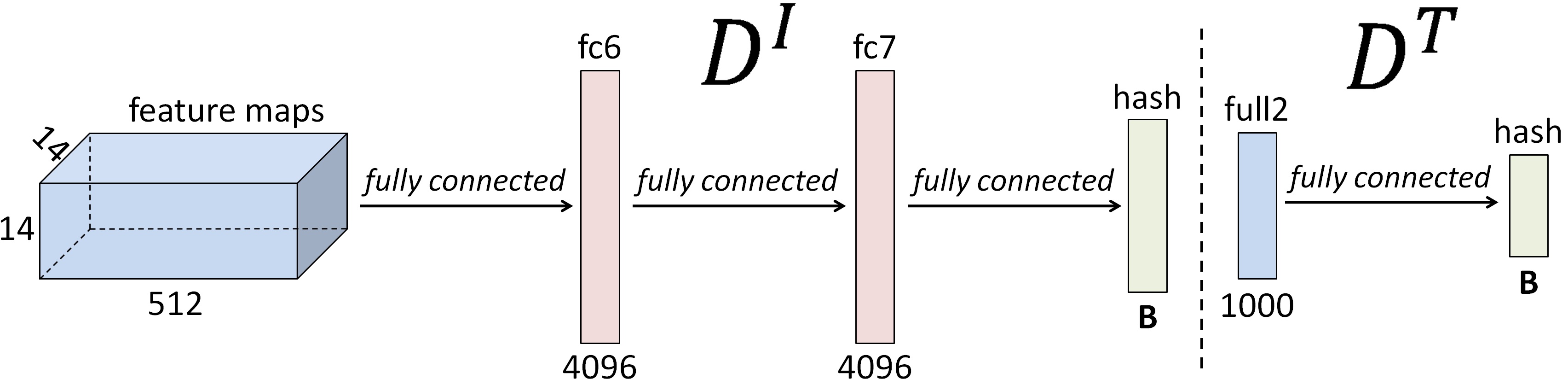}
\setlength{\abovecaptionskip}{5pt}\caption{Discriminative hashing networks for image modality $D^I$ and tex modality $D^T$.}
\label{figure:feature}
\end{figure}

Since we adopt VGGNet as our basic architecture, we simply use the last full-connected layers, e.g., fc6 and fc7~\footnote{The last fully-connected layer (e.g., fc8) is removed since it is for classification problem. }, to encode the images into binary codes. And then we add a fully-connected layer with $q$ dimensional features which followed by a tanh layer that restricts the values in the range $[-1,1]$. Let the outputs of image discriminator network are $H^I_i = D^I(\breve{f}^I_i)$ and $\hat{H}^I_i = D^I(\hat{f}^I_i)$ as the binary codes for the $i$-th attention-aware feature maps and inattention-aware feature maps, respectively.

For text modality, we also simply add a fully-connected layer and a tanh layer to encode the text features into $p$ bits. Similar with image discriminator, $H^T_i = D^T(\breve{F}^T_i)$ and $\hat{H}^T_i = D^T(\hat{F}^T_i)$ are denoted as the binary codes for the attention-aware and inattention-aware features, respectively.

\subsection{Hashing Objectives}
Our objective contains two types of terms: 1) cross-modal retrieval loss, which learns to keep the similarities between different modalities of data, 2) adversarial retrieval loss, generating the attention distribution.

\subsubsection{Cross-modal Retrieval Loss}
The aim of cross-modal loss function is to keep the similarities between images and texts. To keep the semantic similarities, inter-modal ranking loss and intra-modal ranking loss are used according to ~\cite{yang2017pairwise}. That is the hash codes from different modalities should preserve semantic similarity, and the hash codes from same modality should also preserve semantic similarity.

The cross-modal retrieval loss can be formulated as
 \begin{equation}
 \min \mathcal{F}_{T \to I} + \mathcal{F}_{I \to T} + \mathcal{F}_{I \to I} + \mathcal{F}_{T \to T}
 \end{equation}
where $A \to B$ is denoted as the $A$ modality is taken as the query to retrieve the relevant data of the $B$ modality where $A = [I,T]$ and $B = [I,T]$. For example, $T \to I$ means text queries are used to retrieve relevant images. We denote $\mathcal{F}$ as the similarity preserving loss, and $\mathcal{F}_{A \to B}$ is the loss function for $A$ modality as query and $B$ modality as database. The first two terms are used to preserve the semantic similarity between different modalities, and the last two terms are to preserve the similarity in their own modality.

We take $\mathcal{F}_{T \to I}$ as an example for illustration. Given a binary code $H^T_i$ of the $i$-th text, good hash functions should require that the similar images should rank ahead of the dissimilar images. That is we should have $||H^T_i - H^I_{j}|| \leq ||H^T_i - H^I_{k} ||$ in Hamming space when $S(i,j) > S(i,k)$. Formally, $\mathcal{F}_{T \to I}$ can be defined as
\begin{equation}
\begin{aligned}
\mathcal{F}_{T \to I}&= \sum_{\langle i, j, k \rangle} \max\{0, \varepsilon + ||H^T_i - H^I_j|| - ||H^T_i - H^I_k||\}  \\
s.t.~~~~~~~&\forall \langle i, j, k \rangle, \text{we have} \ \ S(i,j) > S(i,k).
\end{aligned}
\end{equation}
where $\langle i, j, k \rangle$ is the triplet form. The objective is the triplet ranking loss~\cite{lai2015simultaneous} which show effectiveness in the uni-modal retrieval.

Similar with that, $\mathcal{F}_{I \to T}$ can be defined as
\begin{equation}
\begin{aligned}
\mathcal{F}_{I \to T}&= \sum_{\langle i, j, k \rangle} \max\{0, \varepsilon + ||H^I_i - H^T_j|| - ||H^I_i - H^T_k||\}.
\end{aligned}
\end{equation}
The $\mathcal{F}_{I \to I}$ can be formulated as
\begin{equation}
\begin{aligned}
\mathcal{F}_{I \to I}&= \sum_{\langle i, j, k \rangle} \max\{0, \varepsilon + ||H^I_i - H^I_j|| - ||H^I_i - H^I_k||\},  \\
\end{aligned}
\end{equation}
and $\mathcal{F}_{T \to T}$ is
\begin{equation}
\begin{aligned}
\mathcal{F}_{T \to T}&= \sum_{\langle i, j, k \rangle} \max\{0, \varepsilon + ||H^T_i - H^T_j|| - ||H^T_i - H^T_k||\}.  \\
\end{aligned}
\end{equation}

\subsubsection{Adversarial Retrieval Loss}
Inspired by the impressive results in image generation of the generative adversarial network (GAN), we adopt it for generating the attention distribution. Similar with GAN, our method also has two models: generative attention model $G^I,G^T$ and discriminative hashing model $D^I,D^T$. Models $D$ is to preserve the semantic similarity between different modalities. While $G$ tries to generate attention distribution as described in Subsection~\ref{attention_networks}. The inattention-aware features from $G$ should let $D$ fail to keep the semantic similarities. Hence, the adversarial loss can be expressed as
\begin{equation}
\begin{aligned}
&\mathcal{F}_{T \to \hat{I}} + \mathcal{F}_{I \to \hat{T}} =  & \\
& \sum_{\langle i, j, k \rangle} \max\{0, \varepsilon + ||H^T_i - \hat{H}^I_j|| - ||H^T_i - \hat{H}^I_k||\} \\
& +\sum_{\langle i, j, k \rangle} \max\{0, \varepsilon + ||H^I_i - \hat{H}^T_j|| - ||H^I_i - \hat{H}^T_k||\}
\end{aligned}
\end{equation}
where $\hat{I}$ and $\hat{T}$ are the generated inattention-aware features. Note that $[H^I_i, \hat{H}^I_i] = D^I([\breve{f}^I_i,\hat{f}^I_i]) =  D^I(G(f^I_i))$ and $[H^T_i,\hat{H}^I_i] =  D^T(G^T(f^T_i))$. The $G$ try to maximize the loss and $D$ is to minimize the objective.
\begin{equation}
\min_{D^I,D^T} \max_{G^I,G^T} \mathcal{F}_{T \to \hat{I}} + \mathcal{F}_{I \to \hat{T}}
\end{equation}

\subsubsection{Full Objective}
Our full objective is
\begin{equation}
\begin{aligned}
  &\mathcal{F}(E^I,E^T,G^I,G^T,D^I,D^T)=  \ \mathcal{F}_{T \to \hat{I}} + \mathcal{F}_{I \to \hat{T}} \notag \\
     {} \ \ \ \ \ \ \ \ \ \ \ & + \mathcal{F}_{T \to I} + \mathcal{F}_{I \to T} + \mathcal{F}_{I \to I} + \mathcal{F}_{T \to T} \notag
\end{aligned}
\end{equation}

Similar to GAN, we train our model alternatively. The parameters in $G^I$ and $G^T$ are fixed and other parameters are trainable:
\begin{equation}
\min_{E^I,E^T,D^I,D^T} \mathcal{F}(E^I,E^T,G^I,G^T,D^I,D^T).
\end{equation}
And then $E^I,E^T,D^I,D^T$ are fixed and update the generative attention models:
\begin{equation}
\max_{G^I,G^T} \mathcal{F}_{T \to \hat{I}} + \mathcal{F}_{I \to \hat{T}}.
\end{equation}

\section{Experiments}

\begin{table*}[htb]
\small
\setlength{\abovecaptionskip}{2pt}\caption{Comparison about MAP on two cross modal retrieval tasks w.r.t different lengths of hash codes.}
\centering
\renewcommand{\multirowsetup}{\centering}
\begin{tabular}{IcIcIc|c|cIc|c|cIc|c|cI}
\shline
\multirow{2}{2.2cm}{Task} & \multirow{2}{*}{Methods} & \multicolumn{3}{cI}{IAPR TC-12} & \multicolumn{3}{cI}{MIR-Flickr 25k} & \multicolumn{3}{cI}{NUS-WIDE} \\
\cline{3-11}
 & & 16bits & 32bits & 64bits & 16bits & 32bits & 64bits & 16bits & 32bits & 64bits     \\
\shline
\multirow{7}{2.2cm}{Text Query \\ $\downarrow$ \\ Image Database} & CCA & 0.3493 & 0.3438 & 0.3378 & 0.5742 & 0.5713 & 0.5691 & 0.3731 & 0.3661 & 0.3613 \\
\cline{2-11}
 & CMFH & 0.4168 & 0.4212 & 0.4277 & 0.6365 & 0.6399 & 0.6429 & 0.5031 & 0.5187 & 0.5225 \\
\cline{2-11}
 & SCM & 0.3453 & 0.3410 & 0.3470 & 0.6939 & 0.7012 & 0.7060 & 0.5344 & 0.5412 & 0.5484 \\
\cline{2-11}
 & STMH & 0.3687 & 0.3897 & 0.4044 & 0.6074 & 0.6153 & 0.6217 & 0.4471 & 0.4677 & 0.4780 \\
\cline{2-11}
 & SePH & 0.4423 & 0.4562 & 0.4648 & 0.7216 & 0.7261 & 0.7319 & 0.5983 & 0.6025 & 0.6109 \\
\cline{2-11}
 & DCMH & 0.5185 & 0.5378 & 0.5468 & 0.7827 & 0.7900 & 0.7932 & 0.6389 & 0.6511 & 0.6571 \\
\cline{2-11}
 & \textbf{Ours} & \textbf{0.5358} & \textbf{0.5565} & \textbf{0.5648} & \textbf{0.7922} & \textbf{0.8062} & \textbf{0.8074} & \textbf{0.6708} & \textbf{0.6875} & \textbf{0.6939} \\
\shline
\multirow{7}{2.2cm}{Image Query\\ $\downarrow$ \\ Text Database} & CCA & 0.3422 & 0.3361 & 0.3300 & 0.5719 & 0.5693 & 0.5672 & 0.3742 & 0.3667 & 0.3617  \\
\cline{2-11}
 & CMFH & 0.4189 & 0.4234 & 0.4251 & 0.6377 & 0.6418 & 0.6451 & 0.4900 & 0.5053 & 0.5097 \\
\cline{2-11}
 & SCM & 0.3692 & 0.3666 & 0.3802 & 0.6851 & 0.6921 & 0.7003 & 0.5409 & 0.5485 & 0.5553 \\
\cline{2-11}
 & STMH & 0.3775 & 0.4002 & 0.4130 & 0.6132 & 0.6219 & 0.6274 & 0.4710 & 0.4864 & 0.4942 \\
\cline{2-11}
 & SePH & 0.4442 & 0.4563 & 0.4639 & 0.7123 & 0.7194 & 0.7232 & 0.6037 & 0.6136 & 0.6211 \\
\cline{2-11}
 & DCMH & 0.4526 & 0.4732 & 0.4844 & 0.7410 & 0.7465 & 0.7485 & 0.5903 & 0.6031 & 0.6093 \\
\cline{2-11}
 & \textbf{Ours} & \textbf{0.5293} & \textbf{0.5283} & \textbf{0.5439} & \textbf{0.7563} & \textbf{0.7719} & \textbf{0.7720} & \textbf{0.6300} & \textbf{0.6258} & \textbf{0.6468} \\
\shline
\end{tabular}
\label{table:MAP} 
\end{table*}

In this section, we evaluate the performance of our proposed methods on three datasets and compare it with several stage-of-the-art algorithms.

\subsection{Datasets}
We choose three characteristic public datasets for examination: IAPR TC-12\cite{IAPR-TC12}, MIR-Flickr 25k\cite{MIR-Flickr} and NUS-WIDE\cite{NUS-WIDE}.
\par
IAPR TC-12 is a popular dataset for cross modal retrieval. It consists of 20,000 still natural images which are collected from wide domains, with at least one sentence description for each image. The image-text pairs are multi-label, and 255 concept categories are set as the ground truth labels. In our experiment, we use the whole dataset. For image modality, we use the raw pixels directly, and for each text sample, we convert the sentence descriptions into 2912 dimensional bag-of-words vectors.
\par
MIR-Flickr 25k includes 25,000 multi-label images that are downloaded from the photo-sharing website Flickr.com. The textual descriptions of each image are several words. Each instance holds one or more labels among 24 concept categories. In our experiment, we first get rid of the textual words counting less than 20 times, then the image-text pairs lacking in textual words or labels are deleted from the original dataset. Afterwards, we have 20,015 instances. For image modality, we use raw pixels as before, while 1386 dimensional bag-of-words vectors are used to indicate text points for text modality.
\par
NUS-WIDE is a widely used dataset for cross modal retrieval which consists of 269,648 multi-label images. Just as MIR-Flickr, the textual representation of each image is several associated words as well. There are 81 concept categories provided for evaluation. In our experiment, we choose the image-text pairs that belong to the 21 most frequent labels and 1,000 textual words, and the number of which is up to 195,834 subsequently. For image modality, we still use raw pixels, and 1000 dimensional bag-of-words vectors are used for text modality meanwhile.
\par
In order to establish the training and test sets, we choose 2,000 image-text pairs in IAPR TC-12 and MIR-Flickr datasets randomly as test sets, or in other words, query sets. The rest instances form the retrieval sets. 10,000 random samples are selected from the retrieval set as our training sets. Besides, for NUS-WIDE dataset, we select 2,100 image-text pairs as the test or query set. The rest consists the retrieval set, while 10,500 random instances from the retrieval set become the training set. Table~\ref{table:datasets} shows the number of samples in each set intuitively.
\begin{table}[htb]
\setlength{\abovecaptionskip}{2pt}\caption{The number of samples in each dataset.}
\centering
\small
\begin{tabular}{IcIcIcIcI}
\shline
 & IAPR TC-12 & MIR-Flickr & NUS-WIDE \\
\shline
\#Train & 10000 & 10000 & 10500 \\
\hline
\#Test & 2000 & 2000 & 2100 \\
\hline
\#Retrieval & 18000 & 18015 & 193734 \\
\shline
\end{tabular}
\label{table:datasets}
\end{table}
\subsection{Experimental Settings And Evaluation Measures}

\begin{figure*}[htb]
\centering
\subfigure[Query from text to image task. ($T{\rightarrow}I$)]{
\label{figure:pr-curves:a}
\includegraphics[width=0.9\textwidth]{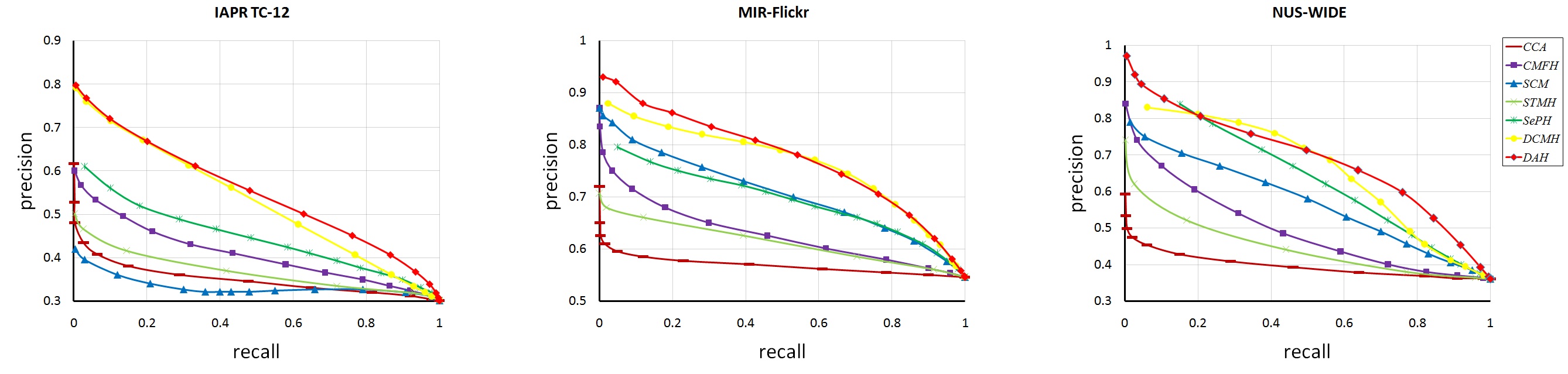}}
\subfigure[Query from image to text task. ($I{\rightarrow}T$)]{
\label{figure:pr-curves:b}
\includegraphics[width=0.9\textwidth]{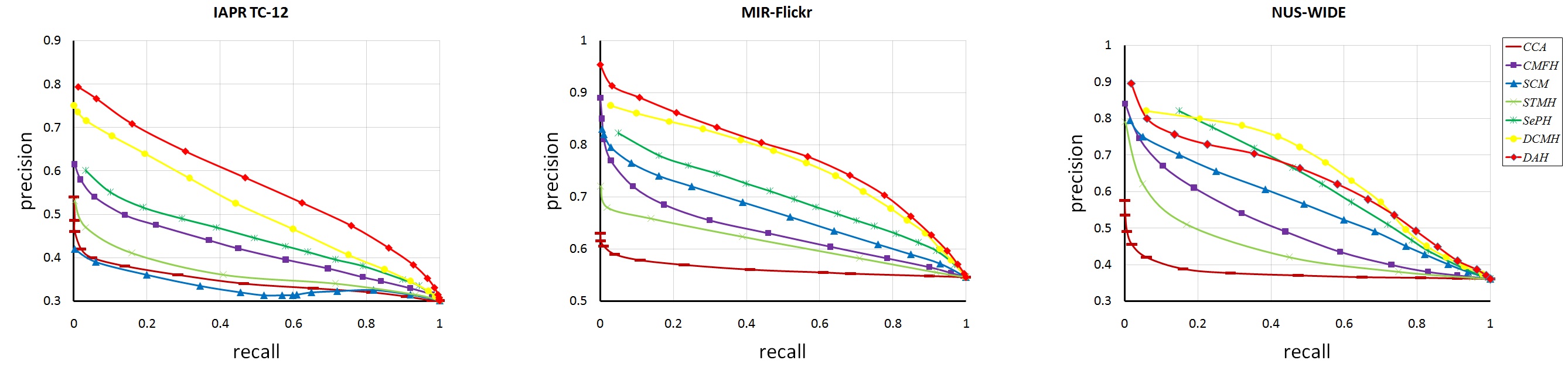}}
\setlength{\abovecaptionskip}{5pt}\caption{precision-recall curves on three datasets. The length of hash code is 16.}
\label{figure:pr-curves}
\end{figure*}

We implement our codes based on the open source \emph{caffe}\cite{caffe} framework. In training, the networks are updated alternatively through the stochastic gradient solver, i.e., ADAM~ ($\alpha=0.0002$, $\beta_{1}=0.5$). We alternate between 4 steps of optimizing $E,D$ and 1 step of optimizing $G$. We initialize the VGGNet on the ImageNet dataset~\cite{ILSVRC} except the last layer. For text modality, all parameters are randomly initialized. The batch size is 64 and the total epoch is 100. The base learning rate is 0.005 and it is changed to one tenth of the current value after every 20 epochs. In testing, we use only the attention-aware features, i.e., foregrounds, of the data to construct the binary codes.

All the samples are ranked according to their Hamming distance from the query. To evaluate the performance of hashing models, we use two metrics: mean average precision (MAP)\cite{hamming} and precision-recall curves. MAP is a standard evaluation metric for information retrieval, which is the mean of averaged precision over a set of queries.

\begin{figure*}[htb]
\centering
\includegraphics[width=0.9\textwidth]{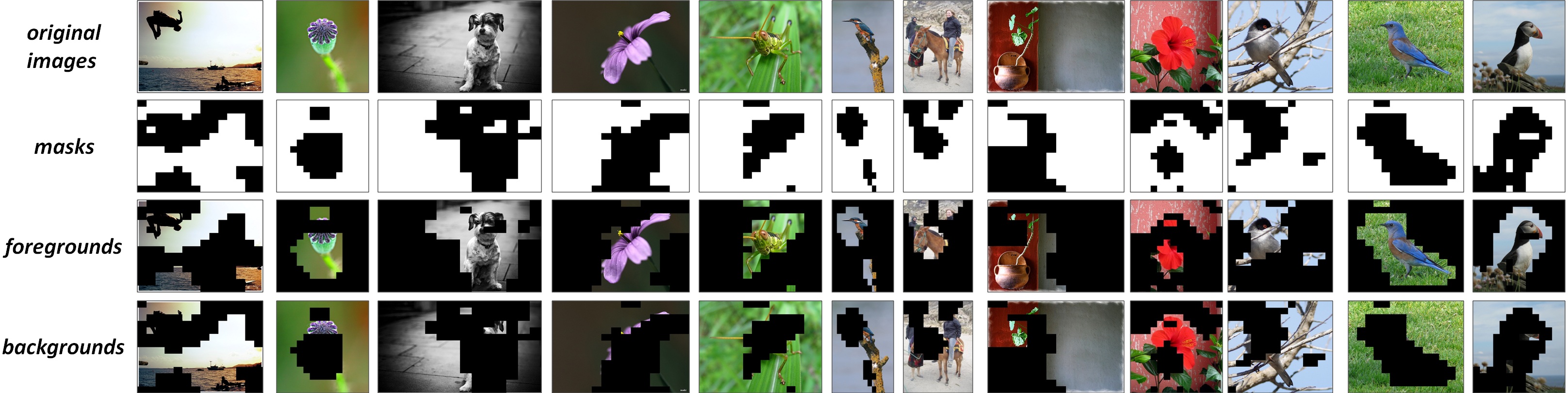}
\setlength{\abovecaptionskip}{5pt}\caption{Some image and mask samples. The first line are original images, The masks are in the middle. The combinations are shown in the bottom.}
\label{figure:maskcompare}
\end{figure*}

\subsection{Comparison with State-of-the-art Methods}
Six state-of-the-art cross-modal hashing approaches are selected as our baselines: CCA\cite{CCA}, CMFH\cite{zhang2014large}, SCM\cite{ding2014collective}, SMTH\cite{SMTH}, SePH\cite{SePH} and DCMH\cite{DCMH}.

The comparison results of search accuracies on all of the three datasets are shown in Table~\ref{table:MAP}. From the table we can see that our method outperforms other baselines and achieves excellent performances. For example, the MAP of our method is 0.5458 compared to 0.5185 of the second best algorithm DCMH. The precision-recall curves are also shown in Figure~\ref{figure:pr-curves}. It can be seen that our method shows comparable performance over the existing baselines.

We also explore the effects of small network architecture in feature learning module for image modality since VGGNet is a large deep network. In this experiment, we select CNN-F~\cite{cnn-f} as the basic network for the image modality. The comparison results are shown in Table~\ref{table:vgg-f}. We can see that VGGNet performs better than CNN-F while our method using CNN-F also achieves good performance when compared to other state-of-the-art baselines.

\begin{table}[htb]
\setlength{\abovecaptionskip}{2pt}\caption{MAP on IAPR TC-12 dataset with different networks.}
\centering
\small
\renewcommand{\multirowsetup}{\centering}
\begin{tabular}{IcIp{1.1cm}<{\centering}Ip{1.1cm}<{\centering}|p{1.1cm}<{\centering}|p{1.1cm}<{\centering}I}
\shline
Task & Network & 16bits & 32bits & 64bits \\
\shline
\multirow{2}{1.1cm}{$T~{\rightarrow}~I$} & VGG & 0.5358 & 0.5565 & 0.5648 \\
\cline{2-5}
 & CNN-F & 0.5267 & 0.5459 & 0.5538 \\
\shline
\hline
\hline
\shline
\multirow{2}{*}{$I~{\rightarrow}~T$} & VGG & 0.5293 & 0.5283 & 0.5439 \\
\cline{2-5}
 & CNN-F & 0.5211 & 0.5168 & 0.5208 \\
\shline
\end{tabular}
\label{table:vgg-f}
\end{table}

The main reason for the good performance of our method is that we can obtain attention distribution for the multi-modal data. Figure~\ref{figure:maskcompare} shows some examples of the image modality. Note that it is hard to visualize the text modality (the networks for text modality use fully-connected layers instead of CNN, and the order of words in BOW are changed), thus we do not show the masks learned in text network.

\section{Conclusion}
\label{conclusions}
In the paper, we proposed a novel approach called HashGAN for the cross-modal hashing based on the idea of adversarial architecture. The proposed HashGAN contains three major components: feature learning module, generative attention module and the discriminative hashing module. The feature learning module learns powerful representations for multi-modal data. The generator and discriminator play two-player minimax game, in which discriminator tries to minimize the similarity-preserving loss functions while generator aims to maximize the retrieval loss of the inattention-aware features. We performed our method on three datasets and the experimental results demonstrate the appealing performance of our method.

\bibliographystyle{ieeetr}
\bibliography{references}
\end{document}